\pdfoutput=1
\documentclass[11pt]{article}
\PassOptionsToPackage{hyperfootnotes=false}{hyperref}
\usepackage[preprint]{acl}
\usepackage{times}
\usepackage{latexsym}
\usepackage[T1]{fontenc}
\usepackage[utf8]{inputenc}
\usepackage{microtype}
\usepackage{graphicx}
\usepackage{amsmath,amssymb}
\usepackage{booktabs,tabularx,array}
\usepackage{multirow}
\usepackage{xurl}
\usepackage{listings}
\newcommand{\fav}{\texttt{favor}}
\newcommand{\agn}{\texttt{against}}
\newcommand{\neu}{\texttt{neutral}}
\newcommand{\PaperAuthors}{%
  Xing Li\textsuperscript{1},
  Jinzhong Ning\textsuperscript{1}\thanks{Corresponding author.},
  Yijia Zhang\textsuperscript{1},
  Liang Yang\textsuperscript{2},
  Hongfei Lin\textsuperscript{2}\\
  \textsuperscript{1}Dalian Maritime University\\
  \textsuperscript{2}Dalian University of Technology\\
  \texttt{lllxxx@dlmu.edu.cn},
  \texttt{ningjinzhong@dlmu.edu.cn}\\
  \texttt{liang@dlut.edu.cn},
  \texttt{hflin@dlut.edu.cn}%
}
\hypersetup{pdftitle={Can Decision Models Understand Stance? Evaluating Jev Against General-Purpose LLMs},
 pdfauthor={Xing Li, Jinzhong Ning, Yijia Zhang, Hongfei Lin},
 pdfsubject={VAST and ZS-CSD stance classification evaluation},
 pdfkeywords={stance detection, Jev, VAST, ZS-CSD, evaluation}}
\title{Can Decision Models Understand Stance? Evaluating Jev Against General-Purpose LLMs}
\author{\PaperAuthors}
\begin{document}
\maketitle
\begin{abstract}
Stance detection requires identifying an author's attitude toward a given target, sometimes based on conversational context. Jev, a specialized decision model designed for structured decision-making, offers an alternative to general-purpose large language models (LLMs). In this work, we evaluate Jev on two stance detection datasets, VAST (English texts) and ZS-CSD (Chinese conversations), comparing it with four general-purpose LLMs and two fine-tuned models. Results show that Jev achieves competitive performance on VAST, matching GPT-5.6 and outperforming the other general-purpose LLMs. However, it falls behind stronger LLMs on ZS-CSD, particularly in distinguishing favor from against. Further analysis suggests that this limitation may be related to understanding reply relationships and stance direction rather than conversation length alone. These findings highlight both the potential and limitations of Jev for stance detection.
\end{abstract}

\section{Introduction}
\label{sec:introduction}

Stance detection aims to determine whether an author is in favor of, against, or neutral toward a given target \citep{mohammad2016}. It is widely used in public opinion analysis \citep{stefanov2020}, rumor detection \citep{wei2019}, and argument mining \citep{barhaim2017}. Unlike sentiment classification, stance detection requires not only identifying the attitude expressed in a text but also determining whether it is directed toward the given target. In practice, authors may express their stance through implicit opinions, indirect evaluations, or disagreement with opposing views, making stance detection more challenging than simply identifying the sentiment of a text. When moving from individual texts to conversations, stance detection also requires understanding the dialogue context \citep{niu2024}. For example, a reply may express its stance by agreeing or disagreeing with an earlier utterance, making it difficult to determine the stance from the reply alone.

In recent years, large language models (LLMs) have provided new ways to address stance detection through their language understanding and contextual modeling abilities \citep{zhang2024lki,weinzierl2024toc}. Unlike conventional approaches that rely on task-specific labeled data, general-purpose LLMs can perform stance detection directly through task instructions, reducing the need for task-specific training \citep{weinzierl2024toc,nguyen2025external}. However, stance detection typically has a fixed set of labels, such as favor, against, and neutral \citep{mohammad2016}. The model only needs to select the appropriate label rather than generate open-ended text. This raises an interesting question: can models designed specifically for structured decision-making achieve performance comparable to general-purpose LLMs on stance detection?

Recently, Jev, a specialized decision model developed by TypeSafe AI, has offered a new way to explore this question \citep{typesafe2026}. Jev is designed to make structured decisions, and its Choice interface selects an answer directly from a predefined set of options. Recent studies have explored its use in scientific, legal, clinical, and rubric-based decision tasks \citep{deng2026,zhang2026,huang2026,rao2026}, with some reporting advantages in efficiency \citep{rao2026}. Although this decision format fits the output requirements of stance detection, it remains unclear whether Jev can accurately identify an author's stance toward a given target, especially in conversations where support or opposition may depend on earlier utterances. Therefore, evaluating Jev across different stance detection settings and comparing it with general-purpose LLMs can help reveal its performance and limitations.

In this work, we conduct an empirical evaluation of Jev on both individual-text and conversational stance detection. Specifically, we use VAST \citep{allaway2020}, an English stance detection dataset, and ZS-CSD \citep{ding2025}, a Chinese zero-shot conversational stance detection dataset, to evaluate Jev in two different settings. We compare Jev with four general-purpose LLMs: DeepSeek V4.1 Flash, Qwen3.8-Flash, GPT-5.6, and Claude Sonnet 5. We also include Qwen2.5-3B and Qwen2.5-7B fine-tuned with LoRA as supervised reference models. Through these experiments, we examine whether Jev can achieve competitive stance detection performance without task-specific training, and analyze its performance differences and error patterns across the two settings.

Our results show that Jev performs differently across the two stance detection settings. On VAST, Jev achieves performance comparable to GPT-5.6 and outperforms the other general-purpose LLMs in our comparison, showing its potential for stance detection on individual texts. However, on ZS-CSD, Jev falls well behind GPT-5.6 and Claude Sonnet 5. Further analysis shows that Jev performs relatively well at distinguishing neutral from non-neutral stances but makes more errors when deciding whether a stance is in favor of or against the target. In addition, the performance gap between Jev and the stronger LLMs does not increase noticeably with conversation length. These findings suggest that Jev's limitations in conversational stance detection may be related to its difficulty in understanding reply relationships and stance direction, rather than conversation length alone. Overall, our study shows that Jev can achieve competitive results in some stance detection settings but still faces challenges when stance judgments depend on conversational context. These findings provide initial evidence of its strengths and limitations in stance detection.

\section{Related Work}
\label{sec:related}

\subsection{Stance detection}

Early work focuses on in-target stance detection, where training and evaluation use the same targets \citep{mohammad2016}. These methods model the relationship between the target and the author's text. For example, \citet{du2017} use target-specific attention to identify relevant parts of the text instead of relying on a target-independent representation.

Cross-target research reduces the need for labeled data for each target by transferring knowledge between related targets. \citet{xu2018} propose CrossNet, using self-attention to extract shared domain features. Zero-shot research evaluates generalization to unseen targets. \citet{allaway2020} introduce VAST and use generalized topic representations to capture similarities between topics. \citet{liang2022} propose JointCL, combining stance contrastive learning with target-aware prototypical graph contrastive learning to transfer stance features to unseen targets.

Recent work also uses LLMs to supply knowledge for stance prediction. \citet{zhang2024lki} extract text--target relationships with an LLM and inject this information into BART. This differs from our direct comparison of Jev with general-purpose LLMs, with fine-tuned models as supervised references.

\subsection{Conversational stance detection}

CSD studies stance expressed through interactions between speakers. Early work such as SRQ distinguishes replies from quoted tweets and evaluates stance prediction for these response types \citep{villacox2020}. Multi-turn conversations require modeling dependencies beyond the immediate reply. \citet{niu2024} introduce MT-CSD and GLAN, which combines global attention with local convolutional and graph-based branches to capture long- and short-range dependencies. These methods use dialogue structure to recover information that may be absent from the current utterance.

Target generalization is also important in conversations, where new discussion topics may lack labeled examples. \citet{ding2025} introduce ZS-CSD, a Chinese benchmark with unseen test targets covering noun phrases and claims. Their SITPCL model combines a speaker interaction network with target-aware prototypical contrastive learning. The interaction network models dependencies within and between speakers, while contrastive learning supports generalization across targets. This line of work brings together two requirements: interpreting the current speaker's stance in context and transferring that interpretation to unfamiliar targets. Our evaluation includes both through the full ZS-CSD test set.

\subsection{Evaluations of Jev}

Recent studies evaluate Jev on scientific \citep{deng2026}, legal \citep{zhang2026}, clinical \citep{huang2026}, and rubric-based \citep{rao2026} judgments. These evaluations concern decisions made under domain-specific criteria. We use stance detection to examine decisions conditioned on a target and, for conversational records, on the history needed to interpret the current utterance.
\section{Method}
\label{sec:method}

We use Jev's Choice interface to solve the tasks in Section~\ref{sec:task}. Each request combines one record's structured input, a task instruction, and a written criterion for each stance label.

\subsection{Task formulation}
\label{sec:task}

\subsubsection{Stance detection}

Given an author's text $u$ and target $t$, the task is to identify the author's expressed stance:
\begin{equation}
 \hat y=f(t,u),\qquad \hat y\in\mathcal{Y},
\end{equation}
where $\mathcal{Y}=\{\text{favor},\text{against},\text{neutral}\}$. The target $t$ may be an entity or a claim. Favor and against indicate support for and opposition to $t$. Neutral follows the benchmark annotations, including cases with no expressed stance or text unrelated to the target \citep{allaway2020,ding2025}.

\subsubsection{Conversational stance detection}

Conversational stance detection uses the same stance label set and adds conversation history to the input. Let $C_n=((s_i,u_i))_{i=1}^{n}$ be a conversation, where $s_i$ identifies the speaker of utterance $u_i$, and let $H_n=((s_i,u_i))_{i=1}^{n-1}$ be the history preceding the current utterance $u_n$. The task is
\begin{equation}
 \hat y_n=f\bigl(t,(s_n,u_n),H_n\bigr),\quad \hat y_n\in\mathcal{Y}.
\end{equation}
The output labels the current utterance $u_n$ toward $t$; history and speaker information provide context for interpreting its stance \citep{ding2025}.

\subsection{Structured input representation}
\label{sec:method-input}

Let $v$ denote the focus item, $H$ its history, and $\tau$ the supplied target type. For VAST, $v=u$ and $H=\tau=\varnothing$. For ZS-CSD, $v=(s_n,u_n)$, $H=H_n$, and $\tau$ indicates a noun-phrase or claim target. We construct the request state as
\begin{equation}
 x=\phi(t,v,H,\tau),
 \label{eq:jev-state}
\end{equation}
where $\phi$ serializes these fields into a structured object. VAST inputs identify the target and focus text. ZS-CSD inputs retain all turns in order with speaker identifiers and explicitly designate the last turn as the focus; one-turn conversations have empty history. We strip surrounding whitespace, preserve the original language, and exclude reference labels and sample-selection metadata.

\subsection{Label-specific selection criteria}
\label{sec:method-criteria}

We associate each label with a fixed written criterion:
\begin{equation}
 \mathcal{O}=\{(y,c_y):y\in\mathcal{Y}\},
 \label{eq:jev-options}
\end{equation}
Here $y$ is the option identifier and $c_y$ its criterion. Favor covers support or approval toward the target, against covers opposition or criticism, and neutral covers no discernible stance, unrelated or informational text, and ambiguity. Both datasets use the same criteria.

The instruction $I$ asks Jev to choose exactly one label for the focus text, using history only to interpret that text. Appendix~\ref{app:prompts} gives the full wording. We supply no in-context demonstrations.

\subsection{Prediction from the Choice response}
\label{sec:method-decision}

Let $\mathcal{J}_{\mathrm{Choice}}$ denote the Jev Choice request and $\pi_{\mathrm{choice}}$ extract its selected option identifier. Our prediction is
\begin{equation}
 \hat y=\pi_{\mathrm{choice}}\!\left(
 \mathcal{J}_{\mathrm{Choice}}(I,x,\mathcal{O})\right),
 \label{eq:jev-prediction}
\end{equation}
with $\hat y\in\mathcal{Y}$. The client validates that the returned \texttt{choice} is a stance label and uses it directly, without confidence filtering or adjustment using returned probabilities. Each request classifies one record without task-specific fine-tuning.
\begin{table*}[!t]
\centering
\small
\setlength{\tabcolsep}{3.5pt}
\begin{tabular}{llrrrrr}
\toprule
\textbf{Dataset} & \textbf{Model / setting} & \textbf{Accuracy} & \textbf{Macro-F1} & \textbf{Input tokens} & \textbf{Output tokens} & \textbf{Cost} \\
\midrule
\multirow{7}{*}[-0.5ex]{\textbf{VAST}} & Jev & 78.14 & 77.92 & 1,763,143 & 114,228 & \$0.0741 \\
 & DeepSeek V4.1 Flash & 74.82 & 73.99 & 865,478 & 18,708 & \$0.2133 \\
 & Qwen3.8-Flash & 74.52 & 73.79 & 763,609 & 28,874 & \$0.1021 \\
 & GPT-5.6 & \textbf{78.51} & \textbf{78.08} & 1,178,659 & 33,439 & \$3.9010 \\
 & Claude Sonnet 5 & 74.82 & 74.51 & 2,279,907 & 251,109 & \$4.7051 \\
\cmidrule(lr){2-7}
 & Qwen2.5-3B-Instruct (SFT)$^\dagger$ & 74.16 & 74.52 & --- & --- & --- \\
 & Qwen2.5-7B-Instruct (SFT)$^\dagger$ & 79.32 & 79.91 & --- & --- & --- \\
\midrule
\multirow{7}{*}[-0.5ex]{\textbf{ZS-CSD}} & Jev & 58.40 & 58.43 & 1,846,931 & 98,192 & \$0.0776 \\
 & DeepSeek V4.1 Flash & 58.28 & 56.89 & 815,813 & 15,920 & \$0.1829 \\
 & Qwen3.8-Flash & 54.61 & 53.16 & 707,481 & 27,236 & \$0.0948 \\
 & GPT-5.6 & \textbf{74.65} & \textbf{73.62} & 1,140,780 & 35,592 & \$3.9230 \\
 & Claude Sonnet 5 & 72.52 & 72.42 & 2,402,197 & 325,790 & \$5.5582 \\
\cmidrule(lr){2-7}
 & Qwen2.5-3B-Instruct (SFT)$^\dagger$ & 55.76 & 56.08 & --- & --- & --- \\
 & Qwen2.5-7B-Instruct (SFT)$^\dagger$ & 60.98 & 60.85 & --- & --- & --- \\
\bottomrule
\end{tabular}
\caption{Performance and resource usage on the full test sets of VAST and ZS-CSD. Bold indicates the best results among prompted models. Dashes denote unavailable measurements. $^\dagger$SFT results are from single runs without per-instance predictions.}
\label{tab:main}
\end{table*}

\section{Experiments}
\label{sec:evaluation}

\subsection{Datasets and evaluation metrics}
\label{sec:data}
\label{sec:metrics}

\textbf{VAST} \citep{allaway2020} contains English text--target pairs for stance detection. We evaluate all 3,006 official test records, including the seen-target (few-shot) and unseen-target (zero-shot) subsets, and retain repeated records.

\textbf{ZS-CSD} \citep{ding2025} evaluates Chinese conversational stance toward unseen targets. We use all 2,584 official test records, covering noun-phrase and claim targets. Exact target strings are disjoint across training, development, and test splits. For context, SITPCL, the best model in the original paper, reaches 43.81\% macro-F1 on this test set.

We report accuracy and three-class macro-F1:
\begin{align}
 F_{1,k}&=\frac{2TP_k}{2TP_k+FP_k+FN_k},\\
 F_{\mathrm{macro}}&=\frac{1}{3}\sum_{k=1}^{3}F_{1,k}.
\end{align}
All five prompted runs per dataset use the same test IDs and reference labels. A failed final response receives no accuracy credit and counts as a false negative for its reference class. Paired target-bootstrap intervals use 5,000 resamples, seed 42, and percentile 95\% bounds without multiple-comparison adjustment. Each sampled target contributes all its records to both models, with 759 target clusters in VAST and 45 in ZS-CSD.

\subsection{Models and inputs}
\label{sec:models}

We call a system \emph{prompted} when it is used without task-specific fine-tuning or in-context demonstrations. All five prompted systems receive the target and the focus text, with speaker-tagged history for ZS-CSD. Jev follows the Choice procedure in Section~\ref{sec:method}, while the LLMs receive the label definitions in a shorter prose instruction (Appendix~\ref{app:prompts}).

\textbf{Jev.} We request jev-latest, returned as jev-1.13.0, and classify one record per Choice request.

\textbf{Flash models.} DeepSeek V4.1 Flash \citep{deepseek2026} is called through the Responses API with a schema-constrained stance field, and Qwen3.8-Flash \citep{qwenflash2026} through an OpenAI-compatible Chat Completions API with JSON output. Both use temperature zero and a 32-token output limit, with thinking disabled.

\textbf{GPT-5.6 and Claude Sonnet 5.} GPT-5.6 \citep{openai2026gpt56} runs through Codex CLI with no tools and reasoning effort set to none. Claude Sonnet 5 \citep{anthropic2026sonnet5} is also evaluated with thinking disabled. Unlike the other three systems, which classify one record per request, GPT-5.6 and Claude Sonnet 5 receive up to 100 and 25 records per request, respectively, and are instructed to classify each record independently.

\textbf{SFT references.} As supervised fine-tuning (SFT) references, we train Qwen2.5-3B-Instruct and 7B-Instruct \citep{qwen2024} with LoRA \citep{hu2021} on each training split, with learning rate $5\times10^{-5}$, five epochs, and seed 42. Only their single-run scores are available, so the paired and subgroup analyses cover the five prompted systems.

\subsection{Experimental settings}
\label{sec:resources}

\suppressfloats[t]
\begin{table}[t]
\centering
\small
\begin{tabular}{llrrr}
\toprule
\textbf{Dataset} & \textbf{Model} & \textbf{Against} & \textbf{Favor} & \textbf{Neutral} \\
\midrule
\multirow{5}{*}{\textbf{VAST}} & Jev & 77.13 & 72.71 & \textbf{83.92} \\
 & DeepSeek & 74.15 & 66.83 & 81.00 \\
 & Qwen & 73.88 & 65.52 & 81.97 \\
 & GPT-5.6 & \textbf{78.25} & \textbf{73.95} & 82.05 \\
 & Claude & 75.35 & 70.26 & 77.91 \\
\midrule
\multirow{5}{*}{\textbf{ZS-CSD}} & Jev & 56.66 & 59.58 & 59.04 \\
 & DeepSeek & 54.60 & 51.18 & \textbf{64.89} \\
 & Qwen & 55.59 & 42.45 & 61.42 \\
 & GPT-5.6 & \textbf{79.49} & \textbf{82.70} & 58.67 \\
 & Claude & 75.78 & 77.50 & 63.99 \\
\bottomrule
\end{tabular}
\caption{Per-class F1 (\%). Bold indicates the best result for each class within each dataset.}
\label{tab:classes}
\end{table}

The single-record runs of Jev, DeepSeek V4.1 Flash, and Qwen3.8-Flash were collected on September 21--24, 2026, with 16 concurrent requests and up to three retries with backoff. GPT-5.6 processes each dataset in sequential batches, and Claude Sonnet 5 resumes from checkpoints, splitting failed batches until all records are recovered. Each record is scored once on its final outcome. Jev, DeepSeek V4.1 Flash, GPT-5.6, and Claude Sonnet 5 return valid final predictions throughout both tests. Qwen3.8-Flash has one VAST request rejected by content screening; its effect on the score is small, but it remains in the full-test denominator. Appendix~\ref{app:runs} summarizes the reasoning settings.

In addition to the scores, Table~\ref{tab:main} lists the input and output tokens recorded for retained successful requests or completed batches, together with their cost in US dollars.
\section{Results}
\label{sec:results}

\begin{figure}[t]
\centering
\includegraphics[width=\columnwidth]{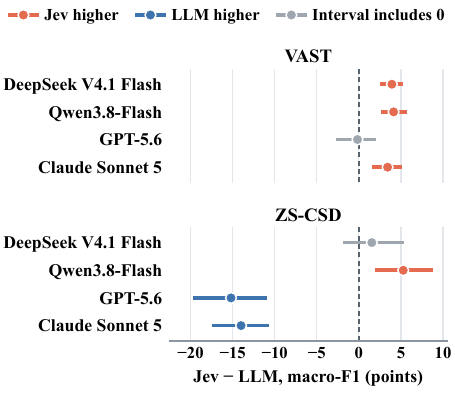}
\caption{Macro-F1 differences between Jev and each LLM (Jev minus LLM, in percentage points), with paired target-bootstrap 95\% intervals.}
\label{fig:paired}
\end{figure}

\begin{table*}[t]
\centering
\small
\setlength{\tabcolsep}{4pt}
\begin{tabular}{lrrrrrr}
\toprule
\textbf{Target type} & $\boldsymbol{N}$ & \textbf{Jev} & \textbf{DeepSeek} & \textbf{Qwen} & \textbf{GPT-5.6} & \textbf{Claude} \\
\midrule
\textbf{Noun phrase} & 967 & 63.49 & 61.61 & 60.80 & \textbf{78.02} & 76.19 \\
\textbf{Claim} & 1,617 & 55.26 & 54.08 & 48.58 & \textbf{70.72} & 70.13 \\
\bottomrule
\end{tabular}
\caption{Macro-F1 scores (\%) across different target types on ZS-CSD. Bold indicates the best result for each target type.}
\label{tab:strata}
\end{table*}

\subsection{Main Results}

Table~\ref{tab:main} presents the overall performance and resource usage of the evaluated models on VAST and ZS-CSD. Figure~\ref{fig:paired} further shows the paired macro-F1 differences between Jev and the four general-purpose LLMs. We summarize the main findings as follows.

\noindent\textbf{(1) Jev performs competitively on VAST but shows a substantial performance gap on ZS-CSD.} On VAST, Jev achieves 77.92\% macro-F1, only 0.16 percentage points below GPT-5.6 (78.08\%) and outperforming the other three general-purpose LLMs. The paired confidence interval for the difference between Jev and GPT-5.6 includes zero, indicating that the observed difference is not statistically significant under our evaluation. However, this competitive performance does not extend to ZS-CSD. Jev achieves 58.43\% macro-F1, falling 15.19 and 13.99 points behind GPT-5.6 and Claude Sonnet 5, respectively. Both differences are supported by the paired confidence intervals. These results suggest that Jev can handle stance classification in individual English texts effectively, but remains less competitive in the more challenging Chinese conversational setting.

\noindent\textbf{(2) Jev offers a favorable performance--cost trade-off, particularly on VAST.} Jev outperforms both DeepSeek V4.1 Flash and Qwen3.8-Flash in macro-F1 on the two datasets, although its advantage over DeepSeek V4.1 Flash on ZS-CSD is not statistically established. More importantly, Jev achieves these results at a relatively low inference cost. On VAST, its recorded API cost is only \$0.0741, compared with \$3.9010 for GPT-5.6, despite their nearly identical macro-F1 scores. Similar cost differences are observed on ZS-CSD, although Jev performs substantially worse than the stronger LLMs. These findings indicate that Jev can be a cost-effective option for stance detection when its accuracy meets task requirements, especially for individual-text classification.

\noindent\textbf{(3) Jev achieves comparable performance to task-specific fine-tuned models without additional training.} On VAST, Jev achieves 77.92\% macro-F1, outperforming the LoRA-tuned Qwen2.5-3B (74.52\%) but falling below Qwen2.5-7B (79.91\%). A similar pattern appears on ZS-CSD, where Jev achieves 58.43\%, compared with 56.08\% and 60.85\% for the two fine-tuned models. Although Jev does not outperform the larger fine-tuned model, it achieves competitive results without requiring task-specific training. This provides an additional reference point for evaluating the potential of specialized decision models on stance detection tasks.

\subsection{Class-level Performance}
\label{sec:classes}

On VAST, Jev achieves performance comparable to GPT-5.6 across all three stance categories (Table~\ref{tab:classes}). GPT-5.6 performs slightly better on favor (73.95\% vs. 72.71\%) and against (78.25\% vs. 77.13\%), while Jev achieves the highest F1 on neutral (83.92\%). These results indicate that Jev maintains competitive performance across different stance categories on individual English texts, consistent with the overall results.

On ZS-CSD, however, the performance differences are more pronounced. GPT-5.6 and Claude Sonnet 5 substantially outperform Jev on favor and against, while their neutral F1 scores remain relatively close. For example, GPT-5.6 exceeds Jev by more than 22 percentage points on both favor and against, but achieves a similar F1 on neutral (58.67\% vs. 59.04\%). The confusion analysis further shows that Jev makes 373 favor--against errors, compared with only 35 for GPT-5.6 and 101 for Claude Sonnet 5 (Table~\ref{tab:confusions}). These findings suggest that Jev's main difficulty on ZS-CSD lies in distinguishing support from opposition, rather than in the neutral category. This may reflect limitations in interpreting stance direction when opinions are expressed through interactions between speakers.

\subsection{Performance across Target Types}
\label{sec:target_types}

Table~\ref{tab:strata} reports the macro-F1 scores on the two target types in ZS-CSD. All five models perform better on noun-phrase targets than on claim targets, suggesting that stance detection becomes more challenging when targets are expressed as complete propositions. For Jev, macro-F1 decreases from 63.49\% on noun phrases to 55.26\% on claims. A similar trend is observed for the other models. This may be because claim targets often contain more complex semantic information, such as negation or qualifying conditions, which requires more precise interpretation of the target.

Jev outperforms the two Flash models on both target types but remains substantially behind GPT-5.6 and Claude Sonnet 5. In particular, the gap between Jev and GPT-5.6 is 14.53 percentage points on noun phrases and 15.46 points on claims. These results indicate that Jev's performance limitations on ZS-CSD are consistent across different target types, rather than being restricted to a particular type of target.

\begin{table}[t]
\centering
\small
\setlength{\tabcolsep}{3pt}
\begin{tabular}{rrrrrrr}
\toprule
\textbf{Turns} & $\boldsymbol{N}$ & \textbf{Jev} & \textbf{DeepSeek} & \textbf{Qwen} & \textbf{GPT-5.6} & \textbf{Claude} \\
\midrule
\textbf{1} & 45 & \textbf{66.67} & 53.55 & 62.69 & 64.34 & 59.80 \\
\textbf{2} & 445 & 56.99 & 52.78 & 51.37 & \textbf{79.68} & 74.77 \\
\textbf{3} & 877 & 57.45 & 54.27 & 50.52 & \textbf{74.95} & 70.44 \\
\textbf{4} & 585 & 58.11 & 57.27 & 50.83 & 69.05 & \textbf{70.77} \\
\textbf{5} & 351 & 54.64 & 57.55 & 52.65 & 68.23 & \textbf{69.97} \\
\textbf{6+} & 281 & 54.00 & 55.80 & 57.92 & 68.20 & \textbf{70.06} \\
\bottomrule
\end{tabular}
\caption{Macro-F1 scores (\%) across different conversation lengths on ZS-CSD. Bold indicates the best result for each group.}
\label{tab:depth}
\end{table}

\subsection{Performance across Conversation Lengths}
\label{sec:length}

Table~\ref{tab:depth} reports the macro-F1 scores across different conversation lengths on ZS-CSD. Jev achieves the highest score on single-turn inputs, although this group contains only 45 samples. For conversations with two or more turns, GPT-5.6 and Claude Sonnet 5 consistently outperform Jev. Notably, the performance gap does not increase steadily with conversation length. For example, the gap between Jev and GPT-5.6 decreases from 22.69 percentage points for two-turn conversations to 14.20 points for conversations with six or more turns. These results suggest that Jev's limitations in conversational stance detection cannot be explained by conversation length alone. Instead, its difficulty may also involve interpreting stance relationships between speakers, even in relatively short conversations.

\bibliography{references}
\appendix
\section{Prompts and Input Construction}
\label{app:prompts}

Inputs retain their original language. Reference labels and sample-selection metadata are not included in model inputs. The instructions and label definitions are given below.

\paragraph{Jev Choice instruction.}
\begin{quote}\small
Determine the stance expressed by the focus text toward the target. Use conversation context only to resolve the focus text. Choose exactly one label according to the criteria.
\end{quote}
The criteria of its three options are:
\begin{itemize}
\item \fav: The focus text supports, approves of, argues for, or expresses a positive attitude toward the target.
\item \agn: The focus text opposes, rejects, criticizes, argues against, or expresses a negative attitude toward the target.
\item \neu: The focus text has no discernible stance toward the target, is unrelated, merely reports information, or is too ambiguous to label as favor or against.
\end{itemize}

\paragraph{Single-record LLM instruction.}
The DeepSeek V4.1 Flash and Qwen3.8-Flash requests begin with this text:
\begin{quote}\small
You are a stance classifier. Determine the stance expressed by the focus text toward the target. Use earlier conversation only to interpret the focus text. Favor means support or approval; against means opposition, rejection, or criticism; neutral means no discernible stance, unrelated, merely informational, or too ambiguous.
\end{quote}
DeepSeek V4.1 Flash constrains the response with a stance-enum schema. Qwen3.8-Flash requests a single-key JSON object, and its label is validated in the client.

\paragraph{Batched LLM instruction.}
GPT-5.6 and Claude Sonnet 5 share the same stance criteria, with explicit instructions to classify each record independently, use earlier turns only to interpret its focus text, avoid tools, and return one prediction per input ID in the same order as JSON. Each batch contains test IDs and inputs, without reference labels.

\paragraph{Input construction.}
Section~\ref{sec:method-input} specifies the structured input construction. We add no translation, retrieval, manual text editing, or few-shot examples.

\section{Reasoning Settings}
\label{app:runs}

Thinking is disabled for all four LLMs.

\section{Pairwise Confusions on ZS-CSD}
\label{app:confusions}

Table~\ref{tab:confusions} counts the ZS-CSD errors between each pair of labels. GPT-5.6 and Claude Sonnet 5 make far fewer favor--against errors than Jev, whereas their errors involving neutral are much closer to Jev's.

\begin{table}[!h]
\centering
\small
\setlength{\tabcolsep}{4pt}
\begin{tabular}{lrrr}
\toprule
\textbf{System} & \textbf{Fav.--Ag.} & \textbf{Fav.--Neu.} & \textbf{Ag.--Neu.} \\
\midrule
Jev & 373 & 304 & 398 \\
DeepSeek V4.1 Flash & 275 & 368 & 435 \\
Qwen3.8-Flash & 462 & \textbf{232} & 479 \\
GPT-5.6 & \textbf{35} & 301 & 319 \\
Claude Sonnet 5 & 101 & 306 & \textbf{303} \\
\bottomrule
\end{tabular}
\caption{Number of ZS-CSD test records whose reference and predicted labels form each pair, in either direction. Bold indicates the fewest errors for each label pair.}
\label{tab:confusions}
\end{table}
\end{document}